\documentclass[11pt]{article}

\usepackage[preprint]{acl}

\usepackage{times}
\usepackage{latexsym}

\usepackage[T1]{fontenc}
\usepackage[utf8]{inputenc}

\usepackage{longtable} % Essential for the longtable environment
\usepackage{array}     % Essential for \arraybackslash and custom column types in tables
\usepackage{ragged2e}  % Essential for \RaggedRight to align text left in table cells
\usepackage{amsmath}   % Often useful for math, assuming you have it already
\usepackage{amsfonts}  % Often useful for math, assuming you have it already
\usepackage{amssymb}   % Often useful for math, assuming you have it already
\usepackage{graphicx}  % If you use any images

\usepackage{microtype}

\usepackage{inconsolata}

\usepackage{graphicx}
\usepackage{float}
\usepackage{amsmath}
\usepackage{amsfonts}
\usepackage{amssymb}
\usepackage{booktabs}
\usepackage{algorithm}
\usepackage{algorithmic}
\usepackage{enumitem} % For custom list environments
\usepackage{xcolor}
\usepackage{colortbl}  % \cellcolor for shading table cells
\definecolor{losscell}{RGB}{250, 233, 228}
\definecolor{gaincell}{RGB}{233, 244, 231}
\usepackage[most]{tcolorbox}
\definecolor{promptblue}{RGB}{0, 102, 204}
\definecolor{lightgray}{RGB}{245, 245, 245}
\definecolor{lightblue}{RGB}{240, 248, 255}
\definecolor{lightgreen}{RGB}{240, 255, 240}
\definecolor{darkgray}{RGB}{64, 64, 64}

\title{The Wording Effect: Quantifying Two-Way Drift in LLM Benchmark Performance}

\author{Shailja Thakur$^1$\thanks{~Corresponding author: \texttt{shailja.thakur@ibm.com}} \quad Sungeun An$^2$ \quad Chad DeLuca$^2$ \quad Hima Patel$^1$ \\
  $^1$IBM Research India \quad $^2$IBM Research Almaden \\}

\begin{document}
\maketitle

\begin{abstract}
A benchmark score comes from a single phrasing of each problem. That single phrasing is treated as if it stood for the whole space of ways the same problem could be asked, but it does not. We show that rephrasing a problem while keeping its meaning and answer fixed routinely flips a model's answer in both directions, so some failures become successes and some successes become failures. We call this \textit{drift}. BenchDrift generates meaning-preserving variations of benchmark problems along four axes, namely linguistic, referential, pragmatic, and structural, and measures how often, and why, correctness flips under each. Across eight models and three benchmarks (GSM8K, MMLU, MATH-Hard), we observe that drift is large in both directions. Two findings stand out. First, phrasing sensitivity does not fade as models get better. Instead, it changes sign. Weak models gain more from rephrasing than they lose, while strong models lose far more than they gain. We find that the best models on a benchmark are therefore the ones whose scores depend most on the wording they happened to be given. Second, the models largely agree on which rephrasings cost the most correct answers even though they differ in how much they drift, so fragility belongs to the rephrasing and not to the model. Furthermore, rephrasing breaks answers a model was confident about, whether the problem is made shorter or longer. Code and Data: \url{https://github.com/IBM/BenchDrift/tree/demo-ui}
\end{abstract}

\section{Introduction}

\begin{figure}[t]
  \centering
  \begin{tcolorbox}[
      enhanced, fontupper=\small,
      colback=lightgray, colframe=darkgray,
      boxrule=0.7pt, arc=2pt,
      title={\small\textbf{One GSM8K problem, four axes, one broken answer}},
      coltitle=white, colbacktitle=darkgray,
      boxsep=1.5pt, left=3pt, right=3pt, top=2pt, bottom=2pt,
      before upper={\setlength{\parskip}{1pt}},
  ]
  \colorbox{lightblue}{\parbox{\dimexpr\linewidth-2\fboxsep}{%
  {\color{promptblue}\textbf{Original}} \hfill ground truth: \textbf{188\,mm}

  Four books are arranged on a shelf. The first book is 31 mm thick while the second book is 50 mm thick. The third book is 5 mm less thick than the second book, and the fourth book is twice as thick as the first book. What is the total thickness of the four books?

  \textit{Phi-4:} 188\,mm \quad\checkmark
  }}

  \vspace{2pt}

  \colorbox{lightgreen}{\parbox{\dimexpr\linewidth-2\fboxsep}{%
  {\color{promptblue}\textbf{Pragmatic}} — artist persona \hfill \textbf{188\,mm} \quad\checkmark

  \textit{In your studio, you have a unique bookshelf display that serves as a piece of art.} The first book \textit{adds a layer of} 31 mm \textit{to your composition}, while the second \textit{contributes} 50 mm\ldots
  }}

  \colorbox{lightgreen}{\parbox{\dimexpr\linewidth-2\fboxsep}{%
  {\color{promptblue}\textbf{Linguistic}} — politeness \hfill \textbf{188\,mm} \quad\checkmark

  \textit{Would you be so kind as to calculate} the total thickness of the four books on the shelf? The first book is 31 mm thick\ldots
  }}

  \colorbox{lightgreen}{\parbox{\dimexpr\linewidth-2\fboxsep}{%
  {\color{promptblue}\textbf{Referential}} — entity substitution \hfill \textbf{188\,mm} \quad\checkmark

  The first book is 31 mm thick while the \textit{next} book is 50 mm thick. The \textit{following} book is 5 mm less thick than the \textit{next} book\ldots
  }}

  \colorbox{lightgray}{\parbox{\dimexpr\linewidth-2\fboxsep}{%
  {\color{promptblue}\textbf{Structural}} — interrogative expansion \hfill \textbf{wrong} \quad$\times$

  \textit{How much less thick is the third book than the second book? What is the thickness of the third book?\ldots What is the total thickness of all four books?}

  \textit{Phi-4:} ``Insufficient information to determine\ldots''
  }}
  \end{tcolorbox}
  \caption{A GSM8K problem reworded along four axes, with Phi-4's answer to each. Rewording never changes the correct answer, 188\,mm, but Phi-4 fails the structural rewording, \textit{replying that there is not enough information}.}
  \label{fig:pipeline-example}
\end{figure}

Benchmark scores come from one phrasing per problem, and models are known to be sensitive to how that phrasing is written~\citep{sclar2024quantifying,ribeiro-etal-2020-beyond,goel2021robustnessgym} --- sensitive enough that which model tops a leaderboard has been shown to change once prompts are optimised, even on the same benchmark~\citep{sadjoli2025optimization}.

What this prior work does not offer is a way to measure this sensitivity that (1) keeps the correct answer fixed across every rephrasing, so a change in correctness can only be explained by the wording, and (2) attributes each change to the specific rephrasing that caused it and to whether it helped or hurt. Robustness and adversarial testing methods surface failures but often do not preserve the ground-truth answer~\citep{goel2021robustnessgym,ribeiro-etal-2020-beyond,wallace2019uat}. Prompt-repair and self-refinement methods improve outcomes without explaining why a given rephrasing helped or hurt~\citep{madaan2023selfrefine,khot2022decomposed,shinn2023reflexion,tian2024debugbench}. Prompt optimisers such as DSPy~\citep{khattab2023dspy} and GEPA~\citep{agrawal2025gepa} \textit{search and discard}: they search for the single best-performing prompt for a task and discard the rest once one is found, so the variation itself leaves no trace in what gets reported.

\begin{figure*}[htbp]
    \centering
    \includegraphics[width=0.75\textwidth]{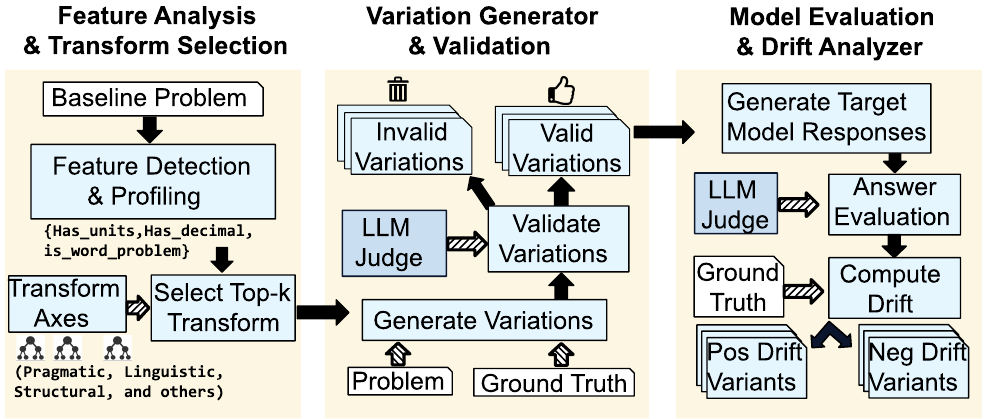}
    \caption{BenchDrift takes a benchmark problem, generates meaning-preserving variations along four axes, validates that each variation still has the same correct answer, and checks whether the model's correctness flips — in either direction.}
    \label{fig:benchdrift_overview}
\end{figure*}

We introduce BenchDrift to close this gap. Given a benchmark problem, BenchDrift generates meaning-preserving variations along four axes — linguistic (wording), referential (which entities are named), pragmatic (tone and context), and structural (sentence organization) — validates that each one keeps the correct answer unchanged (Figure~\ref{fig:pipeline-example} shows one problem reworded along each axis), then collects the model's answers to the original problem and to every valid variation, and compares their correctness (Figure~\ref{fig:benchdrift_overview}). A flip from wrong to right is \textit{positive drift}: a hidden capability the original phrasing was hiding. A flip from right to wrong is \textit{negative drift}: a hidden fragility the original phrasing was masking. Every flip is tied back to the specific rephrasing that produced it.

This paper asks three questions:
\begin{enumerate}[leftmargin=*, itemsep=2pt]
    \item \textbf{How much does correctness change under meaning-preserving rephrasing?} We measure this directly as positive and negative drift rates, for each model and benchmark.
    \item \textbf{Is this change systematic and tied to specific kinds of rephrasing, or is it closer to noise?} We attribute each flip to the transformation that produced it, and test whether different models agree on which transformations cost the most correct answers.
    \item \textbf{What does this mean for how a benchmark score should be read?} We report the best- and worst-case accuracy a model shows under different phrasings, alongside the single-phrasing score currently reported.
\end{enumerate}

\textbf{Contributions:} (1) A meaning-preserving variation framework across four axes, with every variation validated against the original ground truth, and a bidirectional drift measurement over a common denominator that attributes each flip to the transformation that produced it. (2) Two tests that rule out the leading alternative explanations for drift: that rephrasing merely simplifies problems, and that it only disturbs answers the model was unsure of. (3) Reliability checks that swap the generator, validator, and judge models. (4) Open release of the generated and validated variations used in our experiments. (5) Open release of the BenchDrift pipeline and full variation taxonomy.

\section{Method}
\label{sec:method}

\subsection{Roles}
\label{sec:roles}

BenchDrift runs four roles over each problem.

\textit{Generator.} Proposes candidate rephrasings of the problem.

\textit{Validator.} Checks each candidate against the original ground-truth answer and discards any that no longer have it. A changed answer would otherwise mean the problem changed rather than its phrasing.

\textit{Target model.} The model under evaluation. It answers the original problem and every candidate that survives validation.

\textit{Judge.} Scores those answers against the ground truth, accepting ``12'', ``twelve'', and ``a dozen'' as one answer.

Only the target model is under study. The other three are instruments, and all three are language models, so we re-run the pipeline with different models in those roles to check that the drift we report is not an artifact of the instruments themselves (Section~\ref{sec:reliability}).

\subsection{Drift}
\label{sec:drift}

Let $q$ be a benchmark problem with ground-truth answer $a$, and let $V(q)$ be its variations that the validator has confirmed still have answer $a$. Write $C(q) \in \{0,1\}$ for whether the judge scores the target model's answer to $q$ as correct. A flip in $C$ between $q$ and some $q' \in V(q)$ is \textit{drift}, in one of two directions.

\textit{Positive drift} ($C(q)=0$, $C(q')=1$). The model fails the original phrasing and succeeds on the variation: a capability the original wording was hiding.

\textit{Negative drift} ($C(q)=1$, $C(q')=0$). The model succeeds on the original and fails the variation: apparent success that rested on a wording choice.

\textit{Rates.} For a problem set of size $N$ we report both directions as fractions of that same $N$:
\begin{align}
\text{Pos.} &= \tfrac{1}{N}\bigl|\{q : C(q)=0,\ \exists q'\!, C(q')=1\}\bigr| \nonumber \\
\text{Neg.} &= \tfrac{1}{N}\bigl|\{q : C(q)=1,\ \exists q'\!, C(q')=0\}\bigr| \nonumber
\end{align}
The shared denominator makes the two directions comparable: each counts problems, out of the same $N$, that moved one way or the other. Every problem falls into exactly one of four cases: \textit{stable-correct} (right on the original and every variation), \textit{stable-incorrect} (wrong on the original and every variation), \textit{negative drift}, or \textit{positive drift}. These four cases cover the problem set with no overlap.

Best- and worst-case accuracy follow directly from this partition. \textit{Worst} counts a problem correct only if the original and every variation are, so it counts only the stable-correct cases. \textit{Best} counts a problem correct if the original or any variation is, so it adds the positive-drift cases to that same count. Writing $\text{Rep.}$ for the reported single-phrasing accuracy:
\begin{equation}
\text{Best} = \text{Rep.} + \text{Pos.}, \qquad \text{Worst} = \text{Rep.} - \text{Neg.} \nonumber
\end{equation}
The reported score sits inside an interval whose upper margin is the positive drift rate and whose lower margin is the negative rate. Both identities are exact, so Best and Worst can always be recomputed directly from Rep., Pos., and Neg.

Two extensions of this framework recur later in the paper. First, where we instead report a rate conditioned on a subset of problems --- the share of baseline failures with at least one positive-drift variation, for example (Section~\ref{sec:budget}) --- the denominator is that subset rather than $N$, and we state this at each use. Second, every drift event is attributed to the transformation that produced it and, through it, to that transformation's axis (Section~\ref{sec:transformations}).

\subsection{Variation Taxonomy}
\label{sec:taxonomy}

Every transformation BenchDrift can apply comes from a fixed taxonomy of four axes. \textbf{Linguistic} transformations change wording and style, such as rephrasing or switching between active and passive voice. \textbf{Referential} transformations change which entities, units, or values are named, such as swapping a name or converting a unit. \textbf{Pragmatic} transformations change tone, framing, and context, such as adding a persona or recasting the problem as a direct question. \textbf{Structural} transformations change how the problem is organised, such as reordering information or adding text irrelevant to the answer. Each axis groups several transformations, and each transformation requires that a problem carry certain detected properties before it can be applied --- a unit conversion needs a unit, an entity substitution needs a name; transformations with no requirement apply to any problem. Table~\ref{tab:taxonomy} lists, for each axis, the transformations it groups and the properties each one requires.

\begin{table}[t]
  \centering
  \scriptsize
  \setlength{\tabcolsep}{3pt}
  \renewcommand{\arraystretch}{1.12}
  \caption{Transformations exercised in our experiments, by axis, with the properties each requires. Detectors mark twenty properties in total; Appendix~\ref{app:properties} gives the full mapping.}
  \label{tab:taxonomy}
  \begin{tabular}{@{}>{\raggedright\arraybackslash}p{0.9cm} >{\raggedright\arraybackslash}p{3.35cm} >{\raggedright\arraybackslash}p{2.7cm}@{}}
  \toprule
  \textbf{Axis} & \textbf{Transformation} & \textbf{Requires} \\
  \midrule
  Linguistic & rephrasing, narrative style, passive/active voice, politeness, grammar correction & --- \\
   & symbolic representation & \texttt{has\_numbers}, \texttt{has\_math\_operations} \\
  \addlinespace[2pt]
  Referential & entity substitution (1-, 2-, 3-way) & \texttt{has\_entities} \\
  \addlinespace[2pt]
  Pragmatic & ten persona framings: artist, athlete, chef, detective, doctor, engineer, farmer, musician, scientist, teacher & --- \\
  \addlinespace[2pt]
  Structural & interrogative expansion, programming form, format variation, format whitespace & --- \\
   & logical form & \texttt{has\_logical\_structure} \\
   & format quotes & \texttt{has\_quotes} \\
   & format case & \texttt{has\_headers} \\
   & positioning of sections & \texttt{has\_sections}, \texttt{is\_long} \\
   & positioning of paragraphs & \texttt{has\_paragraphs}, \texttt{is\_long} \\
   & quality clarity & \texttt{has\_pronouns}, \texttt{is\_long} \\
   & content removal, quality completeness, added ambiguity & \texttt{is\_long} \\
  \bottomrule
  \end{tabular}
\end{table}

\subsection{Selecting Transformations for a Problem}
\label{sec:selection}

Each problem gets its own set of transformations, chosen in three steps. First, pattern-based detectors scan the problem text and record which properties it has: numbers, fractions, decimals, percentages, 12- and 24-hour times, dates, durations, units, named entities, how many numbers and sentences it contains, and whether it takes more than one step. Second, we compare those properties against the taxonomy, where every transformation lists the properties it needs --- a unit conversion needs a unit, an entity substitution needs a name --- and every axis inherits its transformations' requirements. An axis scores the fraction of its requirements the problem meets, and the four axes are ranked by that score. Third, we hand out the problem's variation budget along that ranking, in the proportions 1.0, 0.75, 0.55, and 0.40, so the best-matching axis gets the most variations and the worst-matching still gets one. Within each axis, we score its transformations the same way and take the highest-scoring until that axis's allotment is used up.

The result is a per-problem list of transformations, weighted toward the axes the problem actually supports. Since the proportions are unequal by design, axes end up with different numbers of variations, which is why we treat axis-level rates as descriptive and rest the analysis on individual transformations (Section~\ref{sec:transformations}).

\subsection{Validating the Variations}
\label{sec:validation}

Generation can change the answer as well as the wording, so every candidate variation is checked before any target model sees it. The validator is asked one question: would a reader who never saw the original arrive at the same answer from this variation alone? Changes to wording, framing, or persona pass. Changes to any number, unit, or logical condition that would move the answer do not. ``30 minutes'' and ``half an hour'' are equivalent; ``8 days'' and ``72 hours'' are not, because 8 days is 192 hours. Candidates that fail are discarded and appear in none of the results we report.

What survives is a set of variations per problem, each carrying the original ground-truth answer. Every target model then answers the original problem and all of its surviving variations, the judge scores each answer, and the drift we report is whatever flips between the original and a variation.

\section{Experimental Setup}
\label{sec:setup}

We evaluate BenchDrift across multiple models and benchmarks to test how much drift occurs, whether it concentrates in specific axes, and what it means for reported benchmark scores.

\textit{Benchmarks.} GSM8K~\cite{cobbe2021training} (grade-school math), MMLU~\cite{hendrycks2021measuring} (multi-domain factual questions), and MATH-Hard (harder multi-step math problems), 500 problems sampled per benchmark. These span mathematical reasoning and factual recall, letting us check whether drift patterns generalize across reasoning types.

\textit{Models.} We evaluate eight open-weight models spanning 7B to 34B parameters: Qwen3-8B~\citep{qwen2024}, Mistral-7B~\citep{jiang2023mistral}, Phi-4~\citep{phi42024}, Granite-3.3-8B~\citep{granite2024_3}, GPT-OSS-20B~\citep{gptoss2024}, Granite-34B-Code~\citep{granite2024}, Llama-3.1-8B~\citep{grattafiori2024llama3}, and Qwen2.5-7B~\citep{qwen2024_5}. Granite-34B-Code is a code model rather than a general instruction-tuned model, and it scores near the bottom on these benchmarks.

\textit{Variation generation and validation.} Mistral-Large-Instruct-2411~\citep{mistrallarge2024} is the generator and GPT-OSS-120B~\citep{gptoss2024} is the validator (Sections~\ref{sec:taxonomy}--\ref{sec:validation}). After discarding candidates that fail validation, 20.2 variations remain per problem on average, ranging from 3 to 92; MMLU's multiple-choice format supports more transformations than GSM8K or MATH-Hard, so its per-problem count runs higher.

\begin{table*}[t]
\centering
\small
\caption{Drift across 8 models and 3 benchmarks (500 problems each), models ordered by mean reported accuracy. Rep.\ is pass@1 accuracy on the original phrasing. Best counts a problem correct if at least one of its validated rephrasings is answered correctly; Worst counts it correct only if every one of its validated rephrasings is (Section~\ref{sec:drift}). $\Delta$ is what rephrasing gains minus what it costs, $\Delta = (\text{Best} - \text{Rep.}) - (\text{Rep.} - \text{Worst})$, shaded \colorbox{losscell}{red} (net loss) or \colorbox{gaincell}{green} (net gain). Bold marks the highest-performing result (Best) for each model--benchmark pair. CIs of $\pm$1.8--4.4pp appear in Appendix~\ref{app:cis}.}
\label{tab:main_results}
\setlength{\tabcolsep}{3.4pt}
\renewcommand{\arraystretch}{1.04}
\begin{tabular}{@{}l rrrr | rrrr | rrrr@{}}
\toprule
& \multicolumn{4}{c|}{\textbf{GSM8K}} & \multicolumn{4}{c|}{\textbf{MMLU}} & \multicolumn{4}{c}{\textbf{MATH-Hard}} \\
\cmidrule(lr){2-5} \cmidrule(lr){6-9} \cmidrule(lr){10-13}
\textbf{Model} & Worst & Rep. & Best & $\Delta$ & Worst & Rep. & Best & $\Delta$ & Worst & Rep. & Best & $\Delta$ \\
\midrule
GPT-OSS-20B      & 38.0 & 94.8 & \textbf{99.0} & \cellcolor{losscell}$-$52.6 & 23.4 & 81.2 & \textbf{97.0} & \cellcolor{losscell}$-$42.0 & 16.2 & 61.8 & \textbf{93.6} & \cellcolor{losscell}$-$13.8 \\
Phi-4            & 38.2 & 93.4 & \textbf{99.2} & \cellcolor{losscell}$-$49.4 & 21.4 & 80.6 & \textbf{95.6} & \cellcolor{losscell}$-$44.2 & 15.2 & 58.8 & \textbf{83.6} & \cellcolor{losscell}$-$18.8 \\
Qwen2.5-7B       & 27.6 & 87.6 & \textbf{98.2} & \cellcolor{losscell}$-$49.4 & 15.2 & 68.0 & \textbf{94.8} & \cellcolor{losscell}$-$26.0 &  7.8 & 49.4 & \textbf{82.4} & \cellcolor{losscell}$-$8.6 \\
Granite-3.3-8B   & 15.2 & 86.2 & \textbf{98.2} & \cellcolor{losscell}$-$59.0 & 13.6 & 64.4 & \textbf{95.0} & \cellcolor{losscell}$-$20.2 &  3.4 & 43.8 & \textbf{85.8} & \cellcolor{gaincell}$+$1.6 \\
Llama-3.1-8B     & 18.8 & 82.6 & \textbf{97.4} & \cellcolor{losscell}$-$49.0 & 16.4 & 62.2 & \textbf{92.6} & \cellcolor{losscell}$-$15.4 &  0.8 & 26.0 & \textbf{69.6} & \cellcolor{gaincell}$+$18.4 \\
Qwen3-8B         &  4.4 & 48.0 & \textbf{88.8} & \cellcolor{losscell}$-$2.8  & 21.0 & 72.0 & \textbf{95.4} & \cellcolor{losscell}$-$27.6 &  2.4 & 38.0 & \textbf{75.8} & \cellcolor{gaincell}$+$2.2 \\
Granite-34B-Code &  1.4 & 48.0 & \textbf{88.6} & \cellcolor{losscell}$-$6.0  &  5.8 & 46.4 & \textbf{89.6} & \cellcolor{gaincell}$+$2.6  &  0.0 & 19.4 & \textbf{66.2} & \cellcolor{gaincell}$+$27.4 \\
Mistral-7B       &  2.8 & 48.4 & \textbf{91.2} & \cellcolor{losscell}$-$2.8  & 14.2 & 54.8 & \textbf{92.4} & \cellcolor{losscell}$-$3.0  &  0.0 &  9.0 & \textbf{46.6} & \cellcolor{gaincell}$+$28.6 \\
\bottomrule
\end{tabular}
\end{table*}

\textit{Judging.} Target models answer with chain-of-thought prompting~\cite{wei2022chain} at temperature 0, which removes sampling noise so that the drift we report is not variance across repeated runs of the same prompt. Each phrasing is answered once, so accuracy on the original phrasing (Rep.\ in Table~\ref{tab:main_results}) is pass@1. Llama-3.3-70B-Instruct scores correctness as an LLM judge~\cite{zheng2023judging}, accepting "12", "twelve", and "a dozen" as correct for ground truth "12" rather than requiring an exact string match.

\textit{Main configuration.} Mistral-Large-Instruct-2411 as generator, GPT-OSS-120B as validator, and Llama-3.3-70B-Instruct as judge is our main configuration, and produces every result in this paper unless we state otherwise. Section~\ref{sec:reliability} is the exception: it reports what happens when different models fill these three roles.

\section{Results}
\label{sec:results}

We evaluate BenchDrift across the eight models and three benchmarks described in Section~\ref{sec:setup} to answer the three questions from the introduction. How much correctness changes under rephrasing, and what that means for a reported score, is answered next; whether that change is systematic and tied to specific kinds of rephrasing is answered in Sections~\ref{sec:drift-by-axis} and~\ref{sec:transformations}.

\subsection{The Range Behind a Score}
\label{sec:range}

A benchmark score comes from one phrasing. We test the same problems under every validated variation and find a wide range of accuracy the model shows depending on which phrasing it gets. Averaged across all 24 model-benchmark pairs, the gap between worst-case accuracy (correct under every phrasing) and best-case accuracy (correct under at least one phrasing) is 74.7 percentage points (\textbf{RQ1}). Phi-4 on GSM8K makes this concrete. Its reported score is 93.4\%. That score drops to 38.2\% in the worst case and rises to 99.2\% in the best case, a 61-point window with the reported number sitting near the top.

The direction of the shift depends on how strong the model already is. The asymmetry between the two directions --- negative minus positive drift --- tracks baseline accuracy almost linearly across our 24 pairs (Pearson $r = 0.98$, slope $1.08$). Above a 60\% baseline, negative drift exceeds positive drift in all twelve pairs. Below it, the two directions split evenly, six favor recovery and six favor loss. A weak model has little to lose and something to gain, so rephrasing surfaces knowledge the original wording failed to elicit (Mistral-7B on MATH-Hard: 37.6\% positive drift against 9.0\% negative). A strong model has much to lose, so the same operation mostly costs it correct answers. For the strongest models, a single-phrasing score therefore tends to overstate robustness (\textbf{RQ3}).

Table~\ref{tab:main_results} shows this directly: GPT-OSS-20B and Phi-4, the two strongest models on GSM8K, both drop below 40\% in the worst case despite baselines above 93\%.

\begin{figure*}[t]
  \centering
  \includegraphics[width=0.8\textwidth]{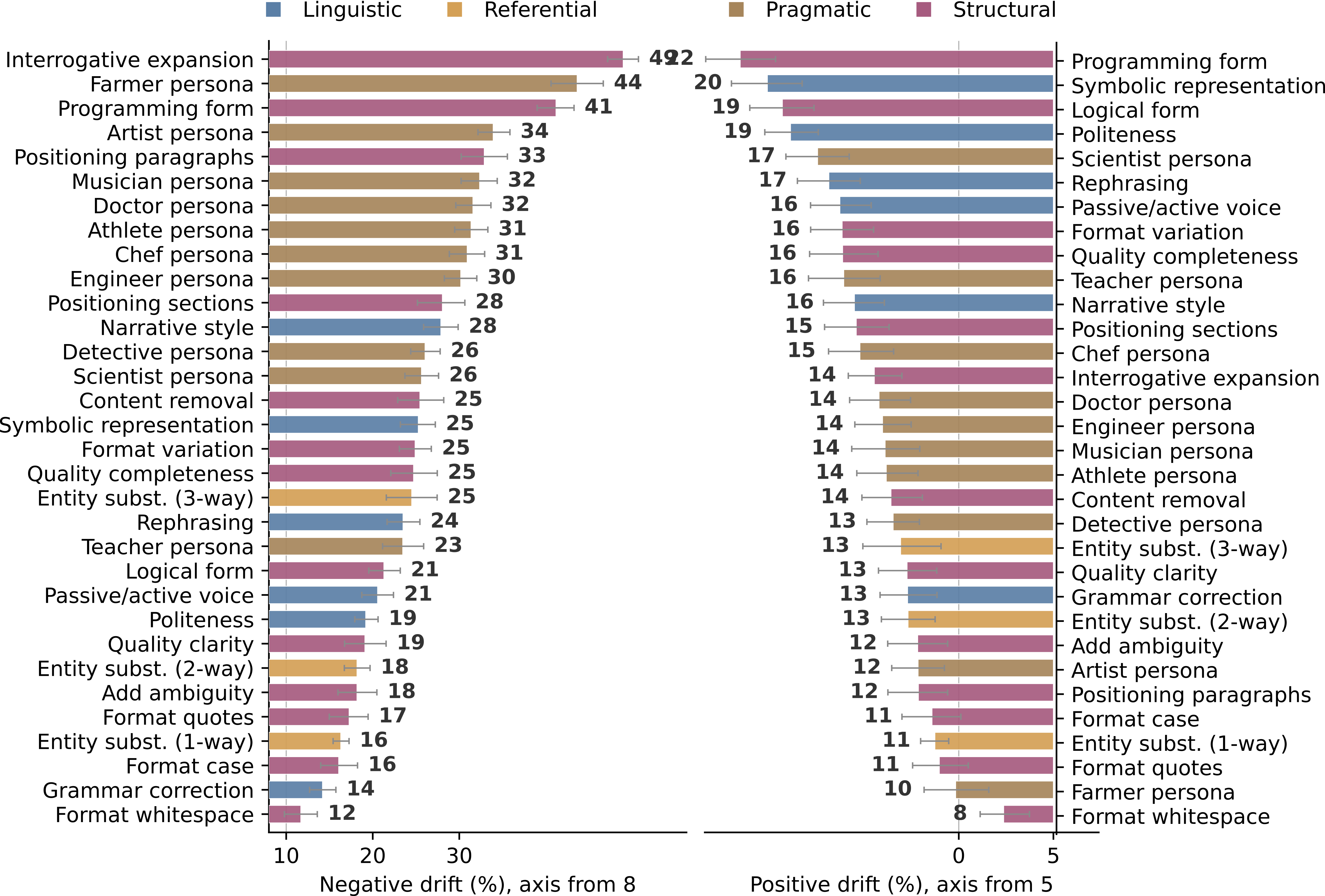}
  \caption{All 32 transformations, pooled over 8 models and 3 benchmarks; colour gives the axis. \textbf{Left:} the share of correct answers a rephrasing broke, ranked by that rate. \textbf{Right:} the share of wrong answers it recovered, ranked independently by that rate. The two panels use different denominators --- answers the model had right, and answers it had wrong --- so they are separate rankings, not parts of one total, and a transformation can rank very differently on each side. Axes start at 8\% and 5\% rather than 0 to keep the bar-length differences legible; printed values give the exact rates. 95\% CIs from a problem-level clustered bootstrap.}
  \label{fig:by_transformation}
\end{figure*}

\begin{figure}[t]
  \centering
  \includegraphics[width=\columnwidth]{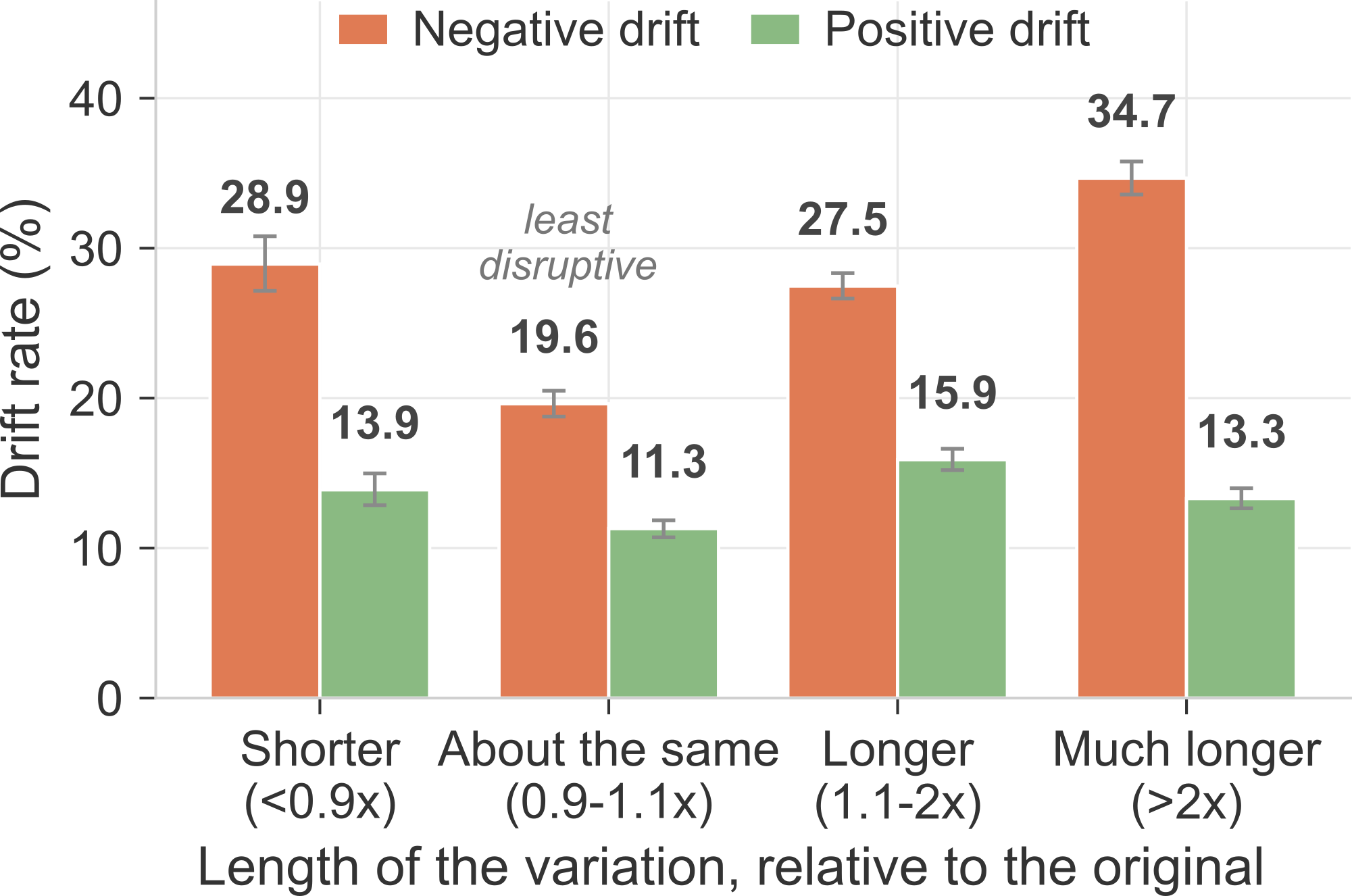}
  \caption{Drift by how much the rephrasing changed the problem's length. Negative drift is lowest when length is preserved and rises in both directions, so drift cannot be attributed to problems simply becoming easier. 95\% CIs from a problem-level clustered bootstrap.}
  \label{fig:by_length}
\end{figure}

\begin{figure}[t]
  \centering
  \includegraphics[width=\columnwidth]{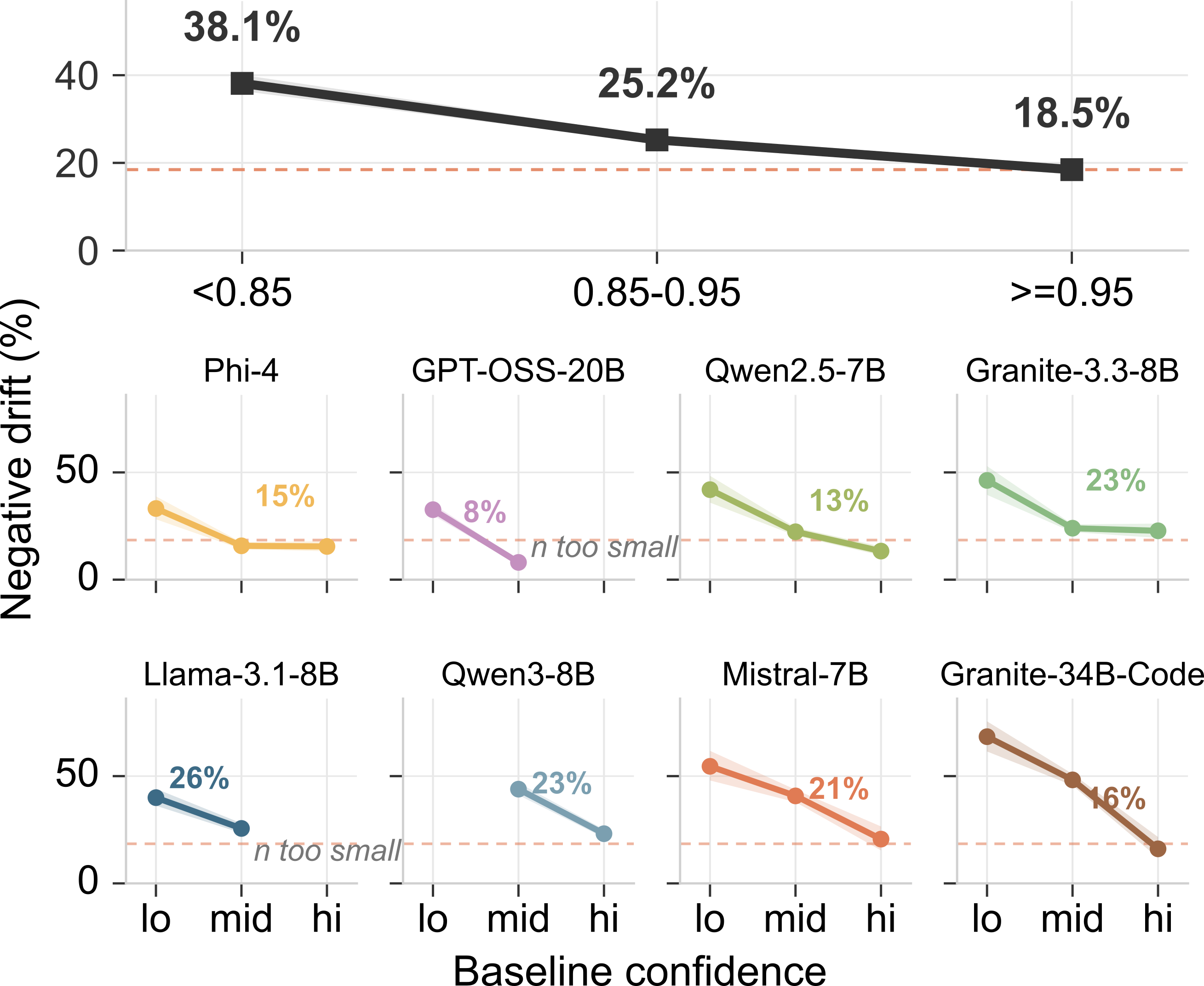}
  \caption{Negative drift by baseline confidence, pooled (top) and per model. Baseline confidence is the mean probability the model assigned to the tokens it generated when answering the original problem, binned as lo ($<$0.85), mid (0.85--0.95), and hi ($\geq$0.95). We use it as a proxy for the model's confidence in its original answer, not as a calibrated estimate. Dashed line marks the pooled hi rate; 95\% CIs from a problem-level clustered bootstrap.}
  \label{fig:by_confidence}
\end{figure}

\subsection{Reliability of the Measurement}
\label{sec:reliability}

Before interpreting this further, we check whether it reflects the models we tested or an artifact of how we measured them. Table~\ref{tab:reliability} reports drift rates and model-fragility rankings under three alternative combinations of generator, validator, and judge model, alongside the main configuration.

\begin{table}[h]
\centering
\scriptsize
\setlength{\tabcolsep}{2.5pt}
\caption{Drift under alternative generator/validator/judge assignments, same 8 models $\times$ 3 benchmarks. Row 1 is the main configuration. $\rho$ is the rank correlation of model fragility against it.}
\label{tab:reliability}
\begin{tabular}{@{}>{\raggedright\arraybackslash}p{1.35cm} >{\raggedright\arraybackslash}p{1.35cm} >{\raggedright\arraybackslash}p{1.35cm} rr r@{}}
\toprule
\textbf{Generator} & \textbf{Validator} & \textbf{Judge} & \textbf{Neg.} & \textbf{Pos.} & \textbf{$\rho$} \\
\midrule
\textbf{Mistral-L.} & \textbf{GPT-OSS-120B} & \textbf{Llama-3.3-70B} & 45.9$\pm$4.3 & 28.8$\pm$3.7 & --- \\
Mistral-L. & Mistral-L. & Mistral-L. & 36.2$\pm$3.9 & 26.0$\pm$3.6 & 0.99 \\
Qwen3-30B & Mistral-L. & Mistral-L. & 34.5$\pm$3.9 & 27.2$\pm$3.7 & 0.98 \\
Qwen3-30B & Qwen3-30B & Qwen3-30B & 34.0$\pm$3.8 & 25.9$\pm$3.6 & 0.98 \\
\bottomrule
\end{tabular}
\end{table}

Swapping every pipeline role shifts average negative drift by at most 11.9 percentage points and average positive drift by at most 2.9, both within the per-cell confidence intervals of Table~\ref{tab:main_results}. The ranking of which models are most fragile is preserved too ($\rho \geq 0.98$ against the main configuration). Two of the configurations differ only in which model generates the variations, so comparing them isolates that role. It moves average negative drift by 1.7 points, so no single role dominates the measurement. Together, these checks support treating the drift we measure as a property of the models under test, not of our measurement pipeline. They do not, however, establish agreement with human judgment, which we have not measured.

\subsection{Drift by Axis}
\label{sec:drift-by-axis}

In Section~\ref{sec:reliability}, we established that drift is a property of the models rather than an artifact of measurement. We next ask whether it spreads evenly across the four axes or concentrates in a few. As Figure~\ref{fig:by_transformation} shows (coloured by axis), it concentrates. We test this by shuffling axis labels within each problem's own variations rather than across problems, since a problem's variations are not independent; the true spread across axes exceeds all 2{,}000 shuffles ($p<0.001$). The axes that break the most answers are the ones that leave a problem's content untouched: \textit{persona} framing and \textit{paragraph reordering} change no number and no relation, yet cost more correct answers than \textit{entity substitution}, which rewrites the tokens the answer depends on. This indicates that the damage tracks surface form, not comprehension.

Because a problem's content determines how much of its transformation budget each axis receives (Section~\ref{sec:taxonomy}), axes with more variations per problem have more chances to flip it, confounding a direct comparison between them. We therefore treat the axis level as descriptive and base the analysis on individual transformations instead.

\subsection{Drift by Transformation}
\label{sec:transformations}

Because axes get unequal shares of a problem's variation budget (Section~\ref{sec:drift-by-axis}), comparing them directly would be unfair, so we rank individual transformations instead. Figure~\ref{fig:by_transformation} ranks all 32 transformations by negative drift and, separately, by positive drift. The negative-drift range alone spans more than fourfold: interrogative expansion, which breaks a problem into a series of smaller explicit questions, costs 48.9\% of correct answers, while adding whitespace costs only 11.7\%. This ranking holds across models too: computing each transformation's negative drift separately for every model and correlating the rankings gives a mean pairwise Spearman $\rho$ of 0.77 across the 28 model pairs. Models differ in how much they drift overall, but largely agree on which transformations cost the most correct answers (\textbf{RQ2}).

Furthermore, negative drift and positive drift are not two sides of the same coin: as Figure~\ref{fig:by_transformation} shows, only \textit{programming formulation} appears in both top-8 lists, causing the most negative drift (41.2\%) and also the most positive drift (21.6\%). Persona and structural transformations otherwise dominate the negative-drift list but cause little positive drift, while linguistic transformations such as \textit{symbolic representation} (25.3\% negative, 20.1\% positive) dominate the positive-drift list without causing much negative drift. Reporting only one direction would miss this.

Colour-coding the transformations by axis shows the axis level still understates fragility: pragmatic and structural transformations cluster at the high-negative-drift end and referential and linguistic ones at the stable end, but the clusters overlap enough that a single transformation crosses lines --- linguistic \textit{narrative style} (27.9\%) breaks far more answers than structural \textit{whitespace} (11.7\%). The transformation, not its axis, predicts fragility.

\subsection{Drift and Problem Length}
\label{sec:not_simplification}

One candidate explanation for why the transformations in Section~\ref{sec:transformations} break answers is that they simply make problems easier or harder. We test this by checking whether drift tracks problem length, using the change in length as a proxy for edit size. As shown in Figure~\ref{fig:by_length}, negative drift is lowest when length is roughly preserved (19.6\%) and rises in \textit{both} directions --- to 28.9\% when the problem gets shorter and 34.7\% when it more than doubles in length. Shortening a problem causes about as much negative drift as lengthening it, which rules out drift as an artifact of problems simply becoming easier.

\subsection{Drift and Model Confidence}
\label{sec:confidence}

A candidate explanation for the drift documented so far is a model's own confidence: low confidence could flag which correct answers are fragile and which scores to distrust. As shown in Figure~\ref{fig:by_confidence}, negative drift falls as confidence rises, from 38.1\% in the low bin to 18.5\% in the high bin, but does not approach zero --- nearly one in five answers the model was most confident about is still lost to a meaning-preserving rephrasing. The pattern holds within every model tested, from 13\% to 26\% in the high bin. Therefore, a model's own confidence cannot be used to tell which of its correct answers will survive rephrasing.

\subsection{How Many Variations Are Needed}
\label{sec:budget}

The results so far use roughly 13 variations per problem to measure drift rates, which is also the cost of running BenchDrift. We now ask how many of those variations are actually needed to see the drift they reveal. We subsample $k$ variations per problem at random and recount how many problems show at least one flip. With the full budget, 67.2\% of problems drift. Five variations recover 74\% of those, eight recover 88\%, and ten recover 93\%; a single variation finds only 30\%.

The curve flattens well before our budget, so an audit on a tighter budget loses less than the reduction in cost suggests. It also does not flatten at one or two variations, which is what a cheap robustness check would use --- most of what we report would be missed at that scale.

\subsection{Comparison to Prompt Optimisers}
\label{sec:dspy_gepa}

In this section, we compare against prior work such as prompt optimisers, which handle phrasing sensitivity differently. DSPy~\citep{khattab2023dspy} and GEPA~\citep{agrawal2025gepa} search over prompts and keep only the best one, a \textit{search-and-discard} strategy, against BenchDrift's \textit{keep-and-measure-the-spread} approach. We run both alongside BenchDrift on the same 15 GSM8K problems, Llama-3.1-8B-Instruct (Table~\ref{tab:dspy_gepa}).

\begin{table}[h]
\centering
\scriptsize
\setlength{\tabcolsep}{4pt}
\caption{Accuracy and wall-clock time per method. DSPy and GEPA each report one optimised prompt's score; BenchDrift reports Worst/Rep./Best across its validated variations.}
\label{tab:dspy_gepa}
\begin{tabular}{@{\extracolsep{\fill}} l r r r r@{}}
\toprule
\textbf{Method} & \textbf{Worst} & \textbf{Rep.} & \textbf{Best} & \textbf{Time} \\
\midrule
DSPy~\citep{khattab2023dspy} & --- & 93.3\% & --- & 1.8s \\
GEPA~\citep{agrawal2025gepa} & --- & 60.0\% & --- & 62.0s \\
BenchDrift & 26.7\% & 86.7\% & 100.0\% & 2.1s \\
\bottomrule
\end{tabular}
\end{table}

At a cost close to DSPy's and far below GEPA's, BenchDrift returns a 73.3-point range (26.7\% to 100\%) that neither optimiser reports, because both search for a single best-performing prompt and discard everything else once they find it. BenchDrift instead keeps the variations its per-problem property detection selects (Section~\ref{sec:selection}), not a random sample of possible rephrasings. On these 15 problems, at comparable cost, DSPy and GEPA hand back a single number that fails to capture a model's real capability, since it offers no way to tell what a different phrasing would have done. BenchDrift instead reports the full range of accuracy across its filtered variations --- worst, representative, and best. Furthermore, it attributes each point in that range to the transformation and axis that produced it.

\section{Related Work}
\label{sec:related}

Models are sensitive to phrasing: formatting alone moves benchmark accuracy by tens of points~\citep{sclar2024quantifying}, equivalent prompts disagree about what a model knows~\citep{elazar-etal-2021-measuring}, and templated edits expose failures a single test set hides~\citep{ribeiro-etal-2020-beyond}. Existing approaches address this sensitivity from different angles. \textbf{Standard benchmarks}~\citep{hendrycks2021measuring,cobbe2021training,wang2019superglue} report accuracy on one phrasing, a single point estimate. \textbf{LLM judges and verifiers}~\citep{zheng2023judging,lightman2023lets} check whether one answer is correct, not whether that correctness holds across equivalent inputs. \textbf{Adversarial and decomposition methods}~\citep{wallace2019uat,goel2021robustnessgym,khot2022decomposed} induce failures by altering the task itself, so a changed answer is uninformative about wording alone.

\textbf{Prompt optimisation and self-refinement methods}~\citep{madaan2023selfrefine,shinn2023reflexion,pryzant2023automatic,khattab2023dspy,agrawal2025gepa} search for a better prompt without isolating which property of the wording mattered. \textbf{Inference-time aggregation}~\citep{wang2022self,jiang2023llm} reduces output variance at 5--40$\times$ cost, but still reports one number. \textbf{Paraphrase and format robustness studies}~\citep{sclar2024quantifying,elazar-etal-2021-measuring,ribeiro-etal-2020-beyond} come closest to our setting, but report that outputs change rather than which kind of change is responsible, without turning the result into a range around the reported score. BenchDrift instead keeps every validated, meaning-preserving variation of a problem and reports the worst, reported, and best accuracy across them, with each point attributable to the specific transformation and axis that produced it.

\section{Conclusion}

A benchmark score computed from one phrasing is not a stable estimate of what a model can do. Across eight models and three benchmarks, accuracy under different valid phrasings of the same problems spans, on average, 74.7 percentage points, and models more often lose correct answers to rephrasing than they gain new ones. We find that drift is systematic. It concentrates in specific transformations, holds up under reliability checks that swap out the entire pipeline, and BenchDrift attributes each flip to the transformation and axis responsible.

\bibliography{references}

\begin{thebibliography}{30}
\providecommand{\natexlab}[1]{#1}

\bibitem[{Abdin et~al.(2024)Abdin, Aneja, Behl, Bubeck, Eldan, Gunasekar,
  Harrison, Hewett, Javaheripi, Kauffmann, Lee, Lee, Li, Liu, Mendes, Nguyen,
  Price, de~Rosa, Saarikivi, Salim, Shah, Wang, Ward, Wu, Yu, Zhang, and
  Zhang}]{phi42024}
Marah Abdin, Jyoti Aneja, Harkirat Behl, Sébastien Bubeck, Ronen Eldan, Suriya
  Gunasekar, Michael Harrison, Russell~J. Hewett, Mojan Javaheripi, Piero
  Kauffmann, James~R. Lee, Yin~Tat Lee, Yuanzhi Li, Weishung Liu, Caio C.~T.
  Mendes, Anh Nguyen, Eric Price, Gustavo de~Rosa, Olli Saarikivi, and others.
  2024.
\newblock \href {https://arxiv.org/abs/2412.08905} {Phi-4 technical report}.
\newblock \emph{Preprint}, arXiv:2412.08905.

\bibitem[{Agarwal et~al.(2025)Agarwal, Ahmad et~al.}]{gptoss2024}
Sandhini Agarwal, Lama Ahmad, and others. 2025.
\newblock \href {https://arxiv.org/abs/2508.10925} {gpt-oss-20b model card}.
\newblock \emph{Preprint}, arXiv:2508.10925.

\bibitem[{Agrawal et~al.(2025)Agrawal, Tan, Soylu, Ziems, Khare, Opsahl-Ong,
  Singhvi, Shandilya, Ryan, Jiang, Potts, Sen, Dimakis, Stoica, Klein, Zaharia,
  and Khattab}]{agrawal2025gepa}
Lakshya~A Agrawal, Shangyin Tan, Dilara Soylu, Noah Ziems, Rishi Khare, Krista
  Opsahl-Ong, Arnav Singhvi, Herumb Shandilya, Michael~J Ryan, Meng Jiang,
  Christopher Potts, Koushik Sen, Alexandros~G Dimakis, Ion Stoica, Dan Klein,
  Matei Zaharia, and Omar Khattab. 2025.
\newblock Gepa: Reflective prompt evolution can outperform reinforcement
  learning.
\newblock \emph{arXiv preprint arXiv:2507.19457}.

\bibitem[{Cobbe et~al.(2021)Cobbe, Kosaraju, Bavarian, Chen, Jun, Kaiser,
  Plappert, Tworek, Hilton, Nakano et~al.}]{cobbe2021training}
Karl Cobbe, Vineet Kosaraju, Mohammad Bavarian, Mark Chen, Heewoo Jun, Lukasz
  Kaiser, Matthias Plappert, Jerry Tworek, Jacob Hilton, Reiichiro Nakano, and
  others. 2021.
\newblock Training verifiers to solve math word problems.
\newblock \emph{arXiv preprint arXiv:2110.14168}.

\bibitem[{Elazar et~al.(2021)Elazar, Kassner, Ravfogel, Ravichander, Hovy,
  Sch{\"u}tze, and Goldberg}]{elazar-etal-2021-measuring}
Yanai Elazar, Nora Kassner, Shauli Ravfogel, Abhilasha Ravichander, Eduard
  Hovy, Hinrich Sch{\"u}tze, and Yoav Goldberg. 2021.
\newblock \href {https://doi.org/10.1162/tacl_a_00410} {Measuring and improving
  consistency in pretrained language models}.
\newblock \emph{Transactions of the Association for Computational Linguistics},
  9:1012--1031.

\bibitem[{Goel et~al.(2021)Goel, Rajani, Vig, Tan, Wu, Zheng, Xiong, Bansal,
  and R{\'e}}]{goel2021robustnessgym}
Karan Goel, Nazneen Rajani, Jesse Vig, Samson Tan, Jason Wu, Stephan Zheng,
  Caiming Xiong, Mohit Bansal, and Christopher R{\'e}. 2021.
\newblock \href {https://arxiv.org/abs/2101.04840} {Robustness gym: Unifying
  the nlp evaluation landscape}.
\newblock \emph{Preprint}, arXiv:2101.04840.

\bibitem[{Granite~Team(2024)}]{granite2024_3}
IBM Granite~Team. 2024.
\newblock \href
  {https://github.com/ibm-granite/granite-3.0-language-models/blob/main/paper.pdf}
  {Granite 3.0 language models}.
\newblock Technical report, IBM.

\bibitem[{Grattafiori et~al.(2024)Grattafiori, Dubey, Jauhri, Pandey, Kadian,
  Al-Dahle, Letman, Mathur, Schelten, Vaughan et~al.}]{grattafiori2024llama3}
Aaron Grattafiori, Abhimanyu Dubey, Abhinav Jauhri, Abhinav Pandey, Abhishek
  Kadian, Ahmad Al-Dahle, Aiesha Letman, Akhil Mathur, Alan Schelten, Alex
  Vaughan, and others. 2024.
\newblock \href {https://arxiv.org/abs/2407.21783} {The llama 3 herd of
  models}.
\newblock \emph{Preprint}, arXiv:2407.21783.

\bibitem[{Hendrycks et~al.(2021)Hendrycks, Burns, Basart, Zou, Mazeika, Song,
  and Steinhardt}]{hendrycks2021measuring}
Dan Hendrycks, Collin Burns, Steven Basart, Andy Zou, Mantas Mazeika, Dawn
  Song, and Jacob Steinhardt. 2021.
\newblock Measuring massive multitask language understanding.
\newblock \emph{Proceedings of ICLR}.

\bibitem[{Jiang et~al.(2023{\natexlab{a}})Jiang, Sablayrolles, Mensch, Bamford,
  Chaplot, Casas, Bressand, Lengyel, Lample, Saulnier
  et~al.}]{jiang2023mistral}
Albert~Q Jiang, Alexandre Sablayrolles, Arthur Mensch, Chris Bamford,
  Devendra~Singh Chaplot, Diego de~las Casas, Florian Bressand, Gianna Lengyel,
  Guillaume Lample, Lucile Saulnier, and others. 2023{\natexlab{a}}.
\newblock Mistral 7b.
\newblock \emph{arXiv preprint arXiv:2310.06825}.

\bibitem[{Jiang et~al.(2023{\natexlab{b}})Jiang, Ren, and Lin}]{jiang2023llm}
Dongfu Jiang, Xiang Ren, and Bill~Yuchen Lin. 2023{\natexlab{b}}.
\newblock Llm-blender: Ensembling large language models with pairwise ranking
  and generative fusion.
\newblock \emph{arXiv preprint arXiv:2306.02561}.

\bibitem[{Khattab et~al.(2023)Khattab, Singhvi, Maheshwari, Zhang, Santhanam,
  Vardhamanan, Haq, Sharma, Joshi, Moazam, Miller, Zaharia, and
  Potts}]{khattab2023dspy}
Omar Khattab, Arnav Singhvi, Paridhi Maheshwari, Zhiyuan Zhang, Keshav
  Santhanam, Sri Vardhamanan, Saiful Haq, Ashutosh Sharma, Thomas~T Joshi,
  Hanna Moazam, Heather Miller, Matei Zaharia, and Christopher Potts. 2023.
\newblock Dspy: Compiling declarative language model calls into self-improving
  pipelines.
\newblock \emph{arXiv preprint arXiv:2310.03714}.

\bibitem[{Khot et~al.(2022)Khot, Trivedi, Finlayson, Fu, Richardson, Clark, and
  Sabharwal}]{khot2022decomposed}
Tushar Khot, Harsh Trivedi, Matthew Finlayson, Yao Fu, Kyle Richardson, Peter
  Clark, and Ashish Sabharwal. 2022.
\newblock Decomposed prompting: A modular approach for solving complex tasks.
\newblock \emph{arXiv preprint arXiv:2210.02406}.

\bibitem[{Lightman et~al.(2023)Lightman, Kosaraju, Burda, Edwards, Baker, Lee,
  Leike, Schulman, Sutskever, and Cobbe}]{lightman2023lets}
Hunter Lightman, Vineet Kosaraju, Yura Burda, Harri Edwards, Bowen Baker, Teddy
  Lee, Jan Leike, John Schulman, Ilya Sutskever, and Karl Cobbe. 2023.
\newblock Let's verify step by step.
\newblock \emph{arXiv preprint arXiv:2305.20050}.

\bibitem[{Madaan et~al.(2023)Madaan, Tandon, Gupta, Hallinan, Gao, Wiegreffe,
  Alon, Dziri, Prabhumoye, Yang, Gupta, Majumder, Hermann, Welleck,
  Yazdanbakhsh, and Clark}]{madaan2023selfrefine}
Aman Madaan, Niket Tandon, Prakhar Gupta, Skyler Hallinan, Luyu Gao, Sarah
  Wiegreffe, Uri Alon, Nouha Dziri, Shrimai Prabhumoye, Yiming Yang, Shashank
  Gupta, Bodhisattwa~Prasad Majumder, Katherine Hermann, Sean Welleck, Amir
  Yazdanbakhsh, and Peter Clark. 2023.
\newblock \href {https://arxiv.org/abs/2303.17651} {Self-refine: Iterative
  refinement with self-feedback}.
\newblock \emph{Preprint}, arXiv:2303.17651.

\bibitem[{Mishra et~al.(2024)Mishra, Stallone et~al.}]{granite2024}
Mayank Mishra, Matt Stallone, and others. 2024.
\newblock \href {https://arxiv.org/abs/2405.04324} {Granite code models: A
  family of open foundation models for code intelligence}.
\newblock \emph{Preprint}, arXiv:2405.04324.

\bibitem[{{Mistral AI}(2024)}]{mistrallarge2024}
{Mistral AI}. 2024.
\newblock \href {https://mistral.ai/news/mistral-large-2407/} {Large enough}.
\newblock Mistral Large 2.

\bibitem[{Pryzant et~al.(2023)Pryzant, Iter, Li, Lee, Zhu, and
  Zeng}]{pryzant2023automatic}
Reid Pryzant, Dan Iter, Jerry Li, Yin~Tat Lee, Chenguang Zhu, and Michael Zeng.
  2023.
\newblock Automatic prompt optimization with "gradient descent" and beam
  search.
\newblock \emph{arXiv preprint arXiv:2305.03495}.

\bibitem[{Ribeiro et~al.(2020)Ribeiro, Wu, Guestrin, and
  Singh}]{ribeiro-etal-2020-beyond}
Marco~Tulio Ribeiro, Tongshuang Wu, Carlos Guestrin, and Sameer Singh. 2020.
\newblock \href {https://doi.org/10.18653/v1/2020.acl-main.442} {Beyond
  accuracy: Behavioral testing of {NLP} models with {C}heck{L}ist}.
\newblock In \emph{Proceedings of the 58th Annual Meeting of the Association
  for Computational Linguistics}, pages 4902--4912. Association for
  Computational Linguistics.

\bibitem[{Sadjoli et~al.(2025)Sadjoli, Siefken, Ghosh, Mai, and
  Dahlmeier}]{sadjoli2025optimization}
Nicholas Sadjoli, Tim Siefken, Atin Ghosh, Yifan Mai, and Daniel Dahlmeier.
  2025.
\newblock Optimization before evaluation: Evaluation with unoptimised prompts
  can be misleading.
\newblock In \emph{Proceedings of the 63rd Annual Meeting of the Association
  for Computational Linguistics: Industry Track}.

\bibitem[{Sclar et~al.(2024)Sclar, Choi, Tsvetkov, and
  Suhr}]{sclar2024quantifying}
Melanie Sclar, Yejin Choi, Yulia Tsvetkov, and Alane Suhr. 2024.
\newblock \href {https://arxiv.org/abs/2310.11324} {Quantifying language
  models' sensitivity to spurious features in prompt design or: How i learned
  to start worrying about prompt formatting}.
\newblock \emph{Preprint}, arXiv:2310.11324.

\bibitem[{Shinn et~al.(2023)Shinn, Cassano, Berman, Gopinath, Narasimhan, and
  Yao}]{shinn2023reflexion}
Noah Shinn, Federico Cassano, Edward Berman, Ashwin Gopinath, Karthik
  Narasimhan, and Shunyu Yao. 2023.
\newblock Reflexion: Language agents with verbal reinforcement learning.
\newblock \emph{arXiv preprint arXiv:2303.11366}.

\bibitem[{Team(2024)}]{qwen2024_5}
Qwen Team. 2024.
\newblock \href {https://arxiv.org/abs/2412.15115} {Qwen2.5 technical report}.
\newblock \emph{Preprint}, arXiv:2412.15115.

\bibitem[{Tian et~al.(2024)Tian, Ye, Qin, Cong, Lin, Pan, Wu, Hui, Liu, Liu,
  and Sun}]{tian2024debugbench}
Runchu Tian, Yining Ye, Yujia Qin, Xin Cong, Yankai Lin, Yinxu Pan, Yesai Wu,
  Haotian Hui, Weichuan Liu, Zhiyuan Liu, and Maosong Sun. 2024.
\newblock Debugbench: Evaluating debugging capability of large language models.
\newblock \emph{arXiv preprint arXiv:2401.04621}.

\bibitem[{Wallace et~al.(2019)Wallace, Feng, Kandpal, Gardner, and
  Singh}]{wallace2019uat}
Eric Wallace, Shi Feng, Nikhil Kandpal, Matt Gardner, and Sameer Singh. 2019.
\newblock \href {https://arxiv.org/abs/1908.07125} {Universal adversarial
  triggers for attacking and analyzing {NLP}}.
\newblock \emph{Preprint}, arXiv:1908.07125.

\bibitem[{Wang et~al.(2019)Wang, Pruksachatkun, Nangia, Singh, Michael, Hill,
  Levy, and Bowman}]{wang2019superglue}
Alex Wang, Yada Pruksachatkun, Nikita Nangia, Amanpreet Singh, Julian Michael,
  Felix Hill, Omer Levy, and Samuel~R Bowman. 2019.
\newblock Superglue: A stickier benchmark for general-purpose language
  understanding systems.
\newblock In \emph{Advances in Neural Information Processing Systems}, pages
  3266--3280.

\bibitem[{Wang et~al.(2022)Wang, Wei, Schuurmans, Le, Chi, Narang, Chowdhery,
  and Zhou}]{wang2022self}
Xuezhi Wang, Jason Wei, Dale Schuurmans, Quoc Le, Ed~Chi, Sharan Narang,
  Aakanksha Chowdhery, and Denny Zhou. 2022.
\newblock Self-consistency improves chain of thought reasoning in language
  models.
\newblock \emph{arXiv preprint arXiv:2203.11171}.

\bibitem[{Wei et~al.(2022)Wei, Wang, Schuurmans, Bosma, Ichter, Xia, Chi, Le,
  and Zhou}]{wei2022chain}
Jason Wei, Xuezhi Wang, Dale Schuurmans, Maarten Bosma, Brian Ichter, Fei Xia,
  Ed~Chi, Quoc Le, and Denny Zhou. 2022.
\newblock Chain-of-thought prompting elicits reasoning in large language
  models.
\newblock \emph{arXiv preprint arXiv:2201.11903}.

\bibitem[{Yang et~al.(2025)Yang, Li, Yang, Zhang, Hui, Zheng, Yu, Gao, Huang,
  Lv, Zheng, Liu, Zhou, Huang, Hu, Ge, Wei, Lin, Tang, Yang, Tu, Zhang, Yang,
  Yang, Zhou, Zhou, Lin, Dang, Bao, Yang, Yu, Deng, Li, Xue, Li, Zhang, Wang,
  Zhu, Men, Gao, Liu, Luo, Li, Tang, Yin, Ren, Wang, Zhang, Ren, Fan, Su,
  Zhang, Zhang, Wan, Liu, Wang, Cui, Zhang, Zhou, and Qiu}]{qwen2024}
An~Yang, Anfeng Li, Baosong Yang, Beichen Zhang, Binyuan Hui, Bo~Zheng, Bowen
  Yu, Chang Gao, Chengen Huang, Chenxu Lv, Chujie Zheng, Dayiheng Liu, Fan
  Zhou, Fei Huang, Feng Hu, Hao Ge, Haoran Wei, Huan Lin, Jialong Tang, and
  others. 2025.
\newblock \href {https://arxiv.org/abs/2505.09388} {Qwen3 technical report}.
\newblock \emph{Preprint}, arXiv:2505.09388.

\bibitem[{Zheng et~al.(2023)Zheng, Chiang, Sheng, Zhuang, Wu, Zhuang, Lin, Li,
  Li, Xing et~al.}]{zheng2023judging}
Lianmin Zheng, Wei-Lin Chiang, Ying Sheng, Siyuan Zhuang, Zhanghao Wu, Yonghao
  Zhuang, Zi~Lin, Zhuohan Li, Dacheng Li, Eric~P. Xing, and others. 2023.
\newblock Judging llm-as-a-judge with mt-bench and chatbot arena.
\newblock \emph{arXiv preprint arXiv:2306.05685}.

\end{thebibliography}

\appendix
\onecolumn

% Adjust float placement parameters for appendix
\setcounter{totalnumber}{10}
\renewcommand{\topfraction}{0.9}
\renewcommand{\bottomfraction}{0.9}
\renewcommand{\textfraction}{0.1}
\renewcommand{\floatpagefraction}{0.8}

\section{Limitations}
\label{sec:limitations}

\textit{No human validation of the judge or validator.} Both the semantic-equivalence validator and the correctness judge are LLMs. We show that swapping which LLM plays these roles changes results only slightly (Section~\ref{sec:reliability}) — but that shows the roles agree with \textit{each other}, not that any of them agree with a human. We have not run a human-annotation study, so we cannot yet rule out a shared blind spot across all the LLMs we tried. A more specific risk is directional, not just generic uncertainty: if the judge is more permissive in one direction than the other (e.g., more willing to accept a borderline answer on a variation than on the original), that alone could inflate or deflate positive and negative drift asymmetrically — which matters because the asymmetry between them (Section~\ref{sec:results}) is one of our central findings. This is the most important open question for trusting the exact numbers we report.

\textit{Three benchmarks, all closed-form.} GSM8K, MMLU, and MATH-Hard all have a single, checkable correct answer. We do not know whether drift behaves the same way on open-ended tasks (summarization, long-form generation) where correctness itself is not binary — this is a natural and, we think, important next step, not something this paper covers.

\textit{Generator and judge model families overlap with target models.} Several of the models we generate variations with or judge responses with (e.g., Mistral-Large, Qwen3-30B) are from the same families as some of the target models we evaluate. Section~\ref{sec:reliability} shows fragility rankings hold up across role swaps, which is reassuring, but does not fully rule out shared family-level biases affecting all configurations similarly.

\textit{Axis membership is inferred, not recorded.} Our logs identify the transformation applied to each variation but not the axis it belongs to, so we recover axis labels by mapping transformation identifiers onto the taxonomy. The mapping is deterministic and applied uniformly, but it is our own, and a different grouping of the same transformations would shift the axis-level numbers. The transformation-level results in Section~\ref{sec:transformations} do not depend on it.

\textit{Cost.} Running BenchDrift end to end — variation generation, validation, and response collection across many models — is more expensive than evaluating a single phrasing once, which makes it better suited to periodic audits than to routine per-run evaluation. Section~\ref{sec:budget} shows that most of the drift we detect is reachable with a fraction of our full budget, but the first run still costs more than a standard evaluation.

\textit{Scope.} BenchDrift measures whether a benchmark score is stable under meaning-preserving rephrasing; it is not a prompt optimizer, and the specific rephrasings that help one problem will not automatically transfer to another. What transfers is the kind of change that tends to help or hurt for a given model and benchmark.

\section{Transformation-to-Property Mapping}
\label{app:properties}

Detectors mark twenty problem properties: \texttt{has\_numbers}, \texttt{has\_number\_words}, \texttt{has\_decimal}, \texttt{has\_fraction}, \texttt{has\_percentage}, \texttt{has\_math\_operations}, \texttt{has\_units}, \texttt{has\_duration}, \texttt{has\_date}, \texttt{has\_time\_12hr}, \texttt{has\_time\_24hr}, \texttt{has\_temporal\_reference}, \texttt{has\_entities}, \texttt{has\_pronouns}, \texttt{has\_quotes}, \texttt{has\_headers}, \texttt{has\_sections}, \texttt{has\_paragraphs}, \texttt{has\_code\_syntax}, and \texttt{is\_long}.

The taxonomy declares transformations beyond those exercised here, and a transformation runs only when a problem carries every property it requires. On the referential axis this gating is what reduced the axis to entity substitution: the eleven other referential transformations convert times, dates, durations, fractions, decimals, number words, and units, and our three benchmarks rarely contain them. Table~\ref{tab:full_mapping} gives the complete mapping.

Entity substitution is labelled by how many places in the problem the substituted entity occurs: an entity mentioned three times, changed everywhere it occurs, is a 3-way substitution. Changing only some of its mentions would leave a variation whose answer is no longer the same, since a later sentence may still refer back to the original entity, so every occurrence moves together. Depending on how often the substituted entity recurs, this ranges from 1-way up to several dozen.

\begin{table}[h]
\centering
\small
\setlength{\tabcolsep}{4pt}
\caption{Every transformation in the taxonomy and the properties it requires. Transformations in \textbf{bold} were exercised in our experiments.}
\label{tab:full_mapping}
\begin{tabular}{@{}>{\raggedright\arraybackslash}p{1.4cm} >{\raggedright\arraybackslash}p{4.5cm} >{\raggedright\arraybackslash}p{4.5cm}@{}}
\toprule
\textbf{Axis} & \textbf{Transformation} & \textbf{Requires} \\
\midrule
Ling. & hypothetical framing & --- \\
 & \textbf{interrogative} & --- \\
 & \textbf{narrative style} & --- \\
 & irrelevant context & --- \\
 & near transfer, far transfer & --- \\
 & \textbf{symbolic representation} & \texttt{has\_numbers}, \texttt{has\_math\_operations} \\
\midrule
Ref. & 12hr $\rightarrow$ 24hr time & \texttt{has\_time\_12hr} \\
 & 24hr $\rightarrow$ 12hr time & \texttt{has\_time\_24hr} \\
 & duration unit conversion & \texttt{has\_duration} \\
 & date format variation & \texttt{has\_date} \\
 & relative time conversion & \texttt{has\_temporal\_reference} \\
 & fraction $\rightarrow$ decimal & \texttt{has\_fraction} \\
 & fraction $\rightarrow$ percentage & \texttt{has\_fraction} \\
 & decimal $\rightarrow$ fraction & \texttt{has\_decimal} \\
 & word $\rightarrow$ number & \texttt{has\_number\_words} \\
 & number $\rightarrow$ word & \texttt{has\_numbers} \\
 & unit conversion & \texttt{has\_units} \\
 & \textbf{entity substitution} & \texttt{has\_entities} \\
\midrule
Prag. & \textbf{formal tone, casual tone} & --- \\
 & \textbf{simplify, elaborate} & --- \\
\midrule
Struct. & \textbf{whitespace add/normalise} & --- \\
 & \textbf{logical formulation} & \texttt{has\_logical\_structure} \\
 & quote single $\leftrightarrow$ double & \texttt{has\_quotes} \\
 & header case upper/lower & \texttt{has\_headers} \\
 & \textbf{section reverse} & \texttt{has\_sections}, \texttt{is\_long} \\
 & \textbf{paragraph reverse} & \texttt{has\_paragraphs}, \texttt{is\_long} \\
 & examples first, drop examples & \texttt{has\_examples}, \texttt{is\_long} \\
 & \textbf{improve clarity} & \texttt{is\_long} \\
 & \textbf{resolve ambiguity} & \texttt{has\_pronouns}, \texttt{is\_long} \\
 & \textbf{add/remove redundancy} & \texttt{is\_long} \\
\bottomrule
\end{tabular}
\end{table}

\section{Infrastructure and Reproducibility}
\label{app:infra}

\textit{Infrastructure.} All models are served with vLLM on 2 NVIDIA H100 GPUs. Target models answer at temperature 0 (Section~\ref{sec:setup}); the variation generator and validator run at a nonzero temperature to encourage diverse rephrasings, with the exact value given in the released configuration files.

\textit{Code and data release.} The full BenchDrift pipeline, prompt templates for the generator, validator, and judge, the variation taxonomy, and the generated and validated variations used in our experiments are available at \url{https://github.com/IBM/BenchDrift/tree/demo-ui}.

\section{Transformation Examples}
\label{app:examples}

The following illustrates BenchDrift variations of one GSM8K problem, answered by Qwen3-8B; it is a single worked example, not a representative sample across benchmarks or models. \checkmark\ marks a variation Qwen3-8B answered correctly, $\times$ marks one it answered incorrectly.

\smallskip
\noindent\textbf{Baseline:} A store sells apples for \$3 each and oranges for \$5 each. If John buys 4 apples and 2 oranges, how much does he spend in total? \textit{Answer: \$22}

\medskip
\noindent\textbf{1. Rephrasing} \textit{(Linguistic)} \quad\checkmark
\begin{quote}
What is the total amount spent when purchasing 4 apples at \$3 apiece and 2 oranges at \$5 apiece?
\end{quote}

\noindent\textbf{2. Narrative Style} \textit{(Linguistic)} \quad\checkmark
\begin{quote}
John headed to the local store one afternoon. Apples were priced at \$3 each and oranges at \$5. He picked up 4 apples and 2 oranges. How much did he pay at the checkout?
\end{quote}

\noindent\textbf{3. Programming Form} \textit{(Structural)} \quad$\times$
\begin{quote}
Given: apple\_price = 3, orange\_price = 5, apple\_qty = 4, orange\_qty = 2. Compute: total\_cost = (apple\_price $\times$ apple\_qty) + (orange\_price $\times$ orange\_qty). What is total\_cost?
\end{quote}

\noindent\textbf{4. Farmer Persona} \textit{(Pragmatic)} \quad$\times$
\begin{quote}
At the farmers market, a grower prices apples at \$3 per fruit and oranges at \$5 per fruit. A customer takes 4 apples and 2 oranges. What is the total bill?
\end{quote}

\noindent\textbf{5. Hypothetical Framing} \textit{(Linguistic)} \quad$\times$
\begin{quote}
Suppose apples cost \$3 each and oranges cost \$5 each. If someone were to buy 4 apples and 2 oranges, what would the total come to?
\end{quote}

\section{Confidence Intervals for Table~\ref{tab:main_results}}
\label{app:cis}

Gained is the positive drift rate and Lost is the negative drift rate (Section~\ref{sec:drift}), each with its 95\% bootstrap confidence interval, resampled at the problem level.

\begin{table}[h]
\centering
\small
\caption{Per-cell 95\% bootstrap confidence intervals for the drift rates in Table~\ref{tab:main_results}, resampled at the problem level.}
\label{tab:cis}
\begin{tabular}{@{}ll rr@{}}
\toprule
\textbf{Model} & \textbf{Bench} & \textbf{Gained} & \textbf{Lost} \\
\midrule
GPT-OSS-20B & GSM8K & 4.2$\pm$1.8 & 56.8$\pm$4.3 \\
GPT-OSS-20B & MMLU & 15.8$\pm$3.2 & 57.8$\pm$4.3 \\
GPT-OSS-20B & MATH-Hard & 31.8$\pm$4.1 & 45.6$\pm$4.3 \\
Phi-4 & GSM8K & 5.8$\pm$2.1 & 55.2$\pm$4.3 \\
Phi-4 & MMLU & 15.0$\pm$3.1 & 59.2$\pm$4.3 \\
Phi-4 & MATH-Hard & 24.8$\pm$3.8 & 43.6$\pm$4.3 \\
Qwen2.5-7B & GSM8K & 10.6$\pm$2.7 & 60.0$\pm$4.3 \\
Qwen2.5-7B & MMLU & 26.8$\pm$3.9 & 52.8$\pm$4.4 \\
Qwen2.5-7B & MATH-Hard & 33.0$\pm$4.1 & 41.6$\pm$4.3 \\
Granite-3.3-8B & GSM8K & 12.0$\pm$2.9 & 71.0$\pm$4.0 \\
Granite-3.3-8B & MMLU & 30.6$\pm$4.0 & 50.8$\pm$4.4 \\
Granite-3.3-8B & MATH-Hard & 42.0$\pm$4.3 & 40.4$\pm$4.3 \\
Llama-3.1-8B & GSM8K & 14.8$\pm$3.1 & 63.8$\pm$4.2 \\
Llama-3.1-8B & MMLU & 30.4$\pm$4.0 & 45.8$\pm$4.4 \\
Llama-3.1-8B & MATH-Hard & 43.6$\pm$4.3 & 25.2$\pm$3.8 \\
Qwen3-8B & GSM8K & 40.8$\pm$4.3 & 43.6$\pm$4.3 \\
Qwen3-8B & MMLU & 23.4$\pm$3.7 & 51.0$\pm$4.4 \\
Qwen3-8B & MATH-Hard & 37.8$\pm$4.2 & 35.6$\pm$4.2 \\
Granite-34B-Code & GSM8K & 40.6$\pm$4.3 & 46.6$\pm$4.4 \\
Granite-34B-Code & MMLU & 43.2$\pm$4.3 & 40.6$\pm$4.3 \\
Granite-34B-Code & MATH-Hard & 46.8$\pm$4.4 & 19.4$\pm$3.5 \\
Mistral-7B & GSM8K & 42.8$\pm$4.3 & 45.6$\pm$4.3 \\
Mistral-7B & MMLU & 37.6$\pm$4.2 & 40.6$\pm$4.3 \\
Mistral-7B & MATH-Hard & 37.6$\pm$4.2 & 9.0$\pm$2.5 \\
\bottomrule
\end{tabular}
\end{table}

\end{document}